\documentclass[conference]{IEEEtran}
\usepackage{cite}
\usepackage{amsmath,amssymb,amsfonts}
\usepackage{algorithmic}
\usepackage{graphicx}
\usepackage{textcomp}
\usepackage{xcolor}
\usepackage{booktabs}
\usepackage{csquotes}
\usepackage[numbers,sort&compress]{natbib}       % For bibliography; use numeric citations
\def\BibTeX{{\rm B\kern-.05em{\sc i\kern-.025em b}\kern-.08em
    T\kern-.1667em\lower.7ex\hbox{E}\kern-.125emX}}

\IEEEoverridecommandlockouts

\begin{document}

\title{You're Pushing My Buttons: Instrumented Learning of Gentle Button Presses}
%\title{Listening for the Click: Bridging the Gap Between Vision and Instrumented Manipulation}

%\author{\IEEEauthorblockN{Raman Talwar}
%\and
%\IEEEauthorblockN{Remko Proesmans}
%\and
%\IEEEauthorblockN{Thomas Lips}
%\and
%\IEEEauthorblockN{Francis wyffels}
%\and
%\IEEEauthorblockN{Andreas Verleysen}
%}

\author{Raman Talwar$^{1}$, Remko Proesmans$^{1}$, Thomas Lips$^{1}$, Andreas Verleysen$^{1}$ and Francis wyffels$^{1}$% <-this % stops a space
\thanks{*This work was supported by Research Foundation Flanders (grant no. 1S15925N), and by euROBIN (grant no. 101070596). }% <-this % stops a space
\thanks{$^{1}$Raman Talwar, Remko Proesmans, Thomas Lips, Andreas Verleysen and Francis wyffels are with the AI and Robotics Lab (IDLab-AIRO), Ghent University---imec, Ghent, Belgium
        {\tt\small raman.talwar@ugent.be}}%
}

\maketitle
\begin{abstract}
Learning contact-rich manipulation is difficult from cameras and proprioception alone because contact events are only partially observed. We test whether training-time instrumentation, i.e., object sensorisation, can improve policy performance without creating deployment-time dependencies.
Specifically, we study button pressing as a testbed and use a microphone fingertip to capture contact-relevant audio. We use an instrumented button-state signal as privileged supervision to fine-tune an audio encoder into a contact event detector. We combine the resulting representation with imitation learning using three strategies, such that the policy only uses vision and audio during inference.
Button press success rates are similar across methods, but instrumentation-guided audio representations consistently reduce contact force. These results support instrumentation as a practical training-time auxiliary objective for learning contact-rich manipulation policies.
\end{abstract}
\vspace{-0.3em}

\section{Introduction and Related Work}\vspace{-0.5em}
Imitation learning (IL) of contact-rich manipulation is difficult from cameras and proprioception alone because contact events are only partially observed.
% Imitation learning (IL) of contact-rich manipulation tasks can be hard without haptic feedback for the teleoperator. 
% Moreover, large amounts of data \textcolor{red}{SOME CITATION} are required for an IL policy to interpret the available modalities of computer vision and tactile sensing and predict effective actions.
%Vision-only policies often fail in this setting because end-effector contact is partially occluded, and visual observations provide limited information about impact dynamics.
Instrumentation, i.e., object sensorisation, has been identified as a useful robot learning tool: it can automate high-quality demonstrations and provide privileged signals to improve policy performance~\cite{proesmans_instrumentation_2025,proesmans_instrumentation_2026,Junge2023,DBLP:journals/corr/abs-1910-07113}. The key research question of instrumentation is: \emph{Can training-time instrumentation improve performace in manipulation policies without creating inference-time dependence?} 
We investigate this question for button pressing, where performance is measured not only by task success but also by contact quality: the robot should use minimal force for a successful button press.
In this paper, the robot is provided with two sensing modalities for the button‑pressing task: a wrist camera for localisation, and a microphone in the fingertip for capturing contact‑related audio. Prior studies demonstrate that audio signals offer valuable cues for contact‑rich manipulation~\cite{liu2024maniwav}. In the next section, we detail the instrumented setup, data-collection pipeline, and multimodal observation/action space used to study this question. We then describe how instrumentation-supervised audio representations are integrated into the policy through the evaluated training strategies.\vspace{-0.3em}

\section{Methods and Materials}
\begin{figure}[]
\centering
\includegraphics[width=\columnwidth,height=0.18\textheight,keepaspectratio]{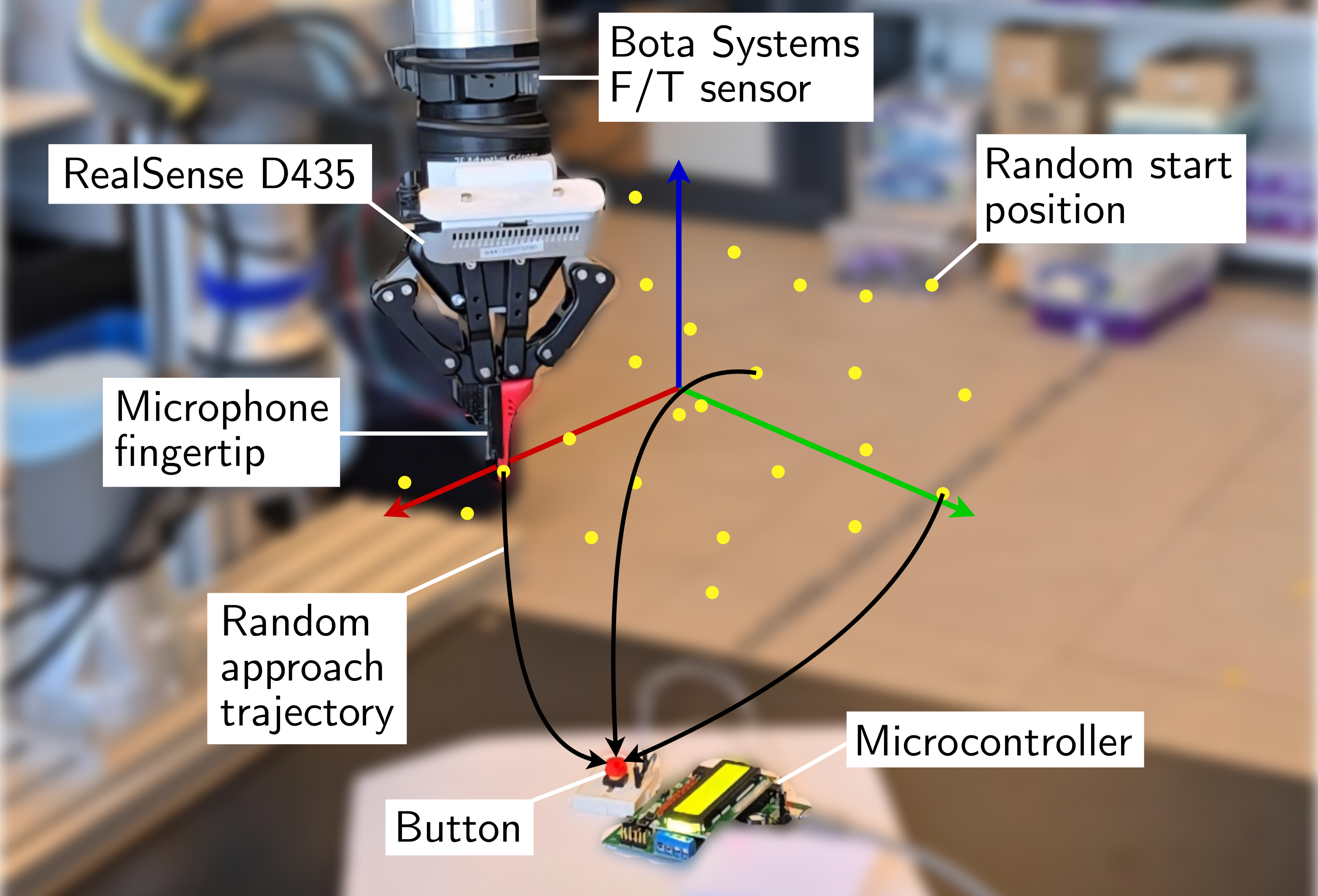}
\caption{Overview of the hardware setup. In addition, some of the randomised end-effector start positions and approach trajectories are indicated.}
\label{fig:materials}
\end{figure}

\vspace{-0.5em}
\textbf{Experimental Setup:}
We use an UR3e arm with a Robotiq~2F-85 gripper, wrist RGB camera (Intel RealSense D435), a Bota~Systems F/T sensor (BFT-DENS-SER-M8), and a custom 3D-printed fingertip with an ICS-43434 microphone (see Fig.~\ref{fig:materials}).
A push button is connected to a microcontroller that reads the binary button state (pressed or unpressed). This button-state signal is \emph{privileged information}: it is available during data collection and training, but not during inference.

\textbf{Task Definition and Automated Data Collection:}
The task is to approach the button, press, and retract. During setup, the button location is measured once and subsequently used as a fixed reference target for all data collection trials. Demonstrations are collected automatically by randomising the end-effector (EEF) start poses and approach trajectories (Fig.~\ref{fig:materials}); once a press event is detected from instrumentation, the robot retracts. This avoids haptic teleoperation and yields high-quality demonstrations~\cite{proesmans_instrumentation_2025}.

\textbf{Observation and Action Spaces:}
For imitation learning, we use Diffusion Policy~\cite{chi2024diffusionpolicy}. To prioritise reactivity, we set the action horizon to 1.
Each observation at time $t$ consists of (policy control at 10\,Hz, i.e., 100\,ms between observations):
\begin{itemize}
    \item \textbf{Wrist RGB image ($\mathbf{I}_t \in \mathbb{R}^{240 \times 320 \times 3}$):} encoded with ResNet-18. Images are randomly cropped to $288 \times 216$ during training, and center-cropped during inference.
    \item \textbf{Audio spectrogram ($\mathbf{A}_t \in \mathbb{R}^{298 \times 128}$):} 3-second log-Mel window.
    \item \textbf{Instrumented Button State ($b_t \in \{0,1\}$):} Binary pressed/unpressed signal, only used in strategies that include privileged instrumentation during training.
\end{itemize}
The robot’s orientation is fixed, and the policy controls only its position. Actions are interpreted as relative displacements expressed in the current EEF frame (i.e., the frame associated with the latest observation).

\textbf{Audio Processing and Audio Encoder Fine-tuning:}
\label{sec:audio_encoder}
To address visual occlusions and the lack of force feedback in image observations, we use the fingertip microphone as an additional (contact)-sensing modality. Audio is processed in 3-second windows, resampled to 16\,kHz, and converted to log-Mel spectrograms. Spectrograms are encoded with an Audio Spectrogram Transformer (AST)~\cite{gong_ast_2021} pretrained on AudioSet~\cite{7952261}. 

To adapt this generic encoder to the task, we fine-tune AST as a binary click detector using labels from the instrumented button signal (10 epochs, learning rate $10^{-5}$). The model reaches F1 $=0.988$ with 1.2\% false negatives on validation, and its representation is used for policy learning.

\begin{figure}[t]
\centering
\includegraphics[width=\columnwidth,height=0.3\textheight,keepaspectratio]{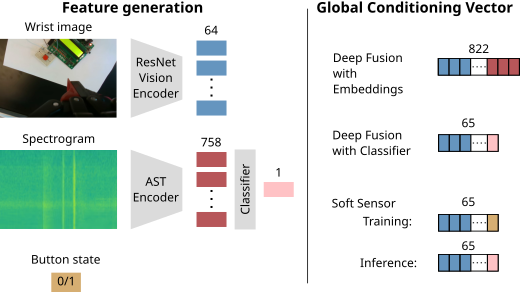}
\caption{Overview of the audio integration strategies evaluated in this work: Deep Fusion (direct conditioning on AST logits or embeddings) and Soft Sensor (training with true button state, replaced by AST prediction at test time).}
\label{fig:integration_strategies}
\end{figure}
\textbf{Integration Strategies:} Fig.~\ref{fig:integration_strategies} summarises the evaluated policy variants:
\begin{itemize}
    \item \textbf{Soft Sensor (Pseudo-Instrumentation):}
    Train the policy with privileged button state in the observation, then replace it at inference with the button-state prediction from the fine-tuned AST click detector.
    \item \textbf{Deep Fusion:} Inject audio during policy training using either (i) the output logits of the fine-tuned AST click detector or (ii) intermediate AST embeddings (penultimate-layer features) from the same fine-tuned model.
\end{itemize}
We compare these variants against a baseline that uses embeddings from a generic AudioSet-pretrained AST without task-specific fine-tuning.

All policies are trained with Adam (learning rate $10^{-4}$, weight decay $10^{-6}$) and a cosine schedule with 500 warmup steps. We apply standard image augmentations (brightness, contrast, saturation, hue, and sharpness jitter). Actions and states are normalised to $[-1,1]$, and spectrograms are normalised following~\cite{gong_ast_2021}.

\textbf{Evaluation Protocol:}
Each policy is evaluated using 40 rollouts. Initial EEF poses are uniformly randomised within the same space used for data collection. A rollout is considered \enquote{successful} if the button is pressed and the robot retracts. Wrist force-torque signals are logged to assess interaction quality, where lower contact force is preferred.\vspace{-0.3em}

\section{Results and Discussion}\vspace{-0.5em}
\begin{figure}[t]
\centering
\includegraphics[width=\columnwidth,height=0.3\textheight,keepaspectratio]{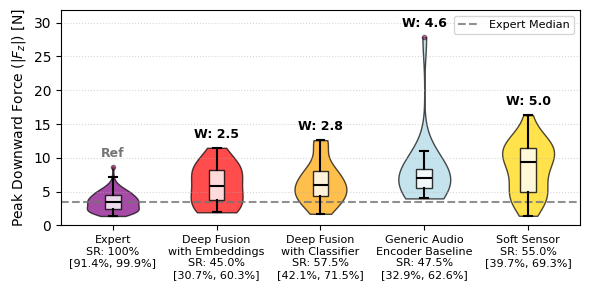}
\caption{Force distributions ranked by Wasserstein distance ($W$ [N]) to the expert demonstrations. Success rates (SR) and Bayesian 95\% credible intervals are provided for each model.}
\label{fig:force_results_workshop}
\end{figure}
%\subsection{Policy Success Rates}
As reported alongside the model labels in Fig.~\ref{fig:force_results_workshop}, success rates fall within a narrow range of 45\% to 55\% across models and are statistically indistinguishable over 40 rollouts per model, with largely overlapping the Bayesian 95\% credible intervals. We therefore focus on contact-force metrics to differentiate policy quality.

%\subsection{Interaction Quality and Force Analysis}
We analysed peak vertical force ($F_z$) during successful rollouts. Compared with expert demonstrations (median $3.41$\,N), the generic Audio Encoder baseline is much harsher (median $6.98$\,N), while Deep Fusion improves contact behavior (median $5.87$\,N with embeddings; $5.95$\,N with classifier). Soft Sensor performs worst (median $9.37N$), likely due to train--test mismatch: the policy trains on perfect button-state signals but receives predicted states at inference, with additional maximum latency of 50\,ms from spectrogram creation.

Computing the Wasserstein distance to the expert force distribution further supports this ranking: Soft Sensor is least similar ($W{=}5.0$\,N), generic AST remains distant ($W{=}4.6$\,N), and Deep Fusion with fine-tuned embeddings ($W{=}2.5$\,N) or fine-tuned classifier ($W{=}2.8$\,N) is closest to expert behavior. These results suggest that conditioning the policy on acoustic representations fine-tuned using instrumentation can effectively regularise contact forces in this task setting.\vspace{-0.3em}

% Computing the Wasserstein distance to the expert force distribution further supports this ranking. We use the Wasserstein distance because it compares empirical force distributions directly, without heavy parametric assumptions, and yields an interpretable metric in Newtons. Soft Sensor is least similar to the expert ($W{=}5.0$\,N), followed by generic AST ($W{=}4.6$\,N), whereas Deep Fusion with fine-tuned embeddings ($W{=}2.5$\,N) or a fine-tuned classifier ($W{=}2.8$\,N) is closest to expert behaviour. These results suggest that conditioning the policy on acoustic representations fine-tuned using instrumentation helps regularise contact forces in this task setting.\vspace{-0.3em}
\section{Conclusion and Future Work}\vspace{-0.5em}
This work suggests that instrumentation can be used during training to improve interaction quality in contact-rich manipulation policies, while inference remains free of privileged sensors. In our experiments, instrumentation-guided audio representation learning does not significantly change success rate, but consistently reduces excessive contact forces.

The main implication is methodological: instrumentation can be treated as a temporary supervision channel for learning deployable representations. While shown here on button pressing, the approach is promising for broader tasks where safe, compliant contact matters, though broader validation across tasks, hardware, and acoustic shifts is still needed.

Next steps include validating this strategy on broader contact-rich tasks (e.g., insertion, latching, and tool use), testing generalisation across different buttons and button types, testing transfer across objects and mechanisms with different contact signatures, and studying tighter integration schemes such as end-to-end audio-visual fusion and partial AST unfreezing during policy training. Together, these directions can help establish instrumentation-guided learning as a general recipe for robust and gentle robotic manipulation.

\bibliography{references.bib}

\end{document}